\documentclass[final,5p,times,twocolumn,numbers]{elsarticle}

\usepackage{amssymb}
\usepackage{lipsum}
\usepackage{amsmath}
\usepackage{multirow}
\usepackage{dblfloatfix} 
\journal{Advanced Engineering Informatic}

\begin{document}

\begin{frontmatter}



\title{Encoding categorical data: Is there yet anything ‘hotter’ than one-hot encoding?}


\author[first]{Poslavskaya Ekaterina}
\affiliation[first]{organization={Huawei Novosibirsk Research Center},%Department and Organization
            %addressline={ Prospect Akademika Lavrentyeva 6/1}, 
            city={Novosibirsk },
            country={Russia}}

\author[first]{Korolev Alexey}

\begin{abstract}
 Categorical features are present in about 40\% of real world problems, highlighting the crucial role of encoding as a preprocessing component. Some recent studies have reported benefits of the various target-based encoders over classical target-agnostic approaches. However, these claims are not supported by any statistical analysis, and are based on a single dataset or a very small and heterogeneous sample of datasets. The present study explores the encoding effects in an exhaustive sample of classification problems from OpenML repository. We fitted linear mixed-effects models to the experimental data, treating task ID as a random effect, and the encoding scheme and the various characteristics of categorical features as fixed effects. We found that in multiclass tasks, one-hot encoding and Helmert contrast coding outperform target-based encoders. In binary tasks, there were no significant differences across the encoding schemes; however, one-hot encoding demonstrated a marginally positive effect on the outcome. Importantly, we found no significant interactions between the encoding schemes and the characteristics of categorical features. This suggests that our findings are generalizable to a wide variety of problems across domains.
\end{abstract}




\end{frontmatter}

%\tableofcontents



\section{Introduction}

Data science seeks to discover patterns observable across multiple tasks. However, in practice, a novel algorithm is often validated on a small sample of datasets, without explanation or justification of the dataset selection procedure, and without proper statistical inference, estimating the significance of the observed effects. This leads to discrepancies in the results across studies and raises questions about generalizability of the reported findings.

In case of  encoding algorithms, any possible advantages of one approach over the other are based on anecdotal or insufficient evidence, coming from a single dataset \citep{Hien2020}, \citep{Potdar2017},  \citep{Bourdonnaye2021}, or from a small number of unbalanced datasets (inter alia \citep{Seca2021}, \citep{Valdez2021}).

Our paper seeks to overcome these limitations for the problem of encoding. We compare different encoding schemes in an exhaustive sample of datasets from a large open source repository OpenML with over 4000 classification tasks \citep{Vanschoren2014}. We report all the stages of how datasets were selected from the original pull of tasks, listing the filtering criteria and the number of tasks excluded at every step. With the resulting sample, we collected experimental data, to which we fitted linear mixed-effects models (LMMs), using task as a random effect. To the best of our knowledge, this is the first paper that explores the effects of encoding, using exhaustive sampling and applying statistical analysis of the aggregated results with LMM to identify possible patterns in the data and the extent of their generalizability.

\subsection{Encoding}

Encoding transforms categorical variables into numbers or vectors of numbers (feature vectors). Few machine learning (ML) models can deal with non-numeric variables directly. First, linear models and deep learning models create linear combinations of dataset features, and string categories cannot be linearly combined. Second, even for tree-based algorithms, the majority of existing libraries require some form of encoding. For example, all category levels must be transformed into a numeric form when using Python scikit-learn library \citep{Pedregosa2012} or XGBoost \citep{ChenGuestrin2016} implementation for R \citealp{Chen2015}. This makes encoding necessary for fitting practically any ML model to categorical data. By our estimates from OpenML, a little bit under 40\% of the real data problems contain categorical features (561 out of 1449 unique classification datasets have at least one categorical variable). This high prevalence of categorical features makes the encoding an essential tool in a data scientist’s toolbox across domains.

Recent studies have reported benefits of some types of encoding over others in specific tasks \citep{Hien2020}, \citep{Potdar2017}, \citep{Seca2021}. However, the conditions making one encoding scheme systematically outperform others are unclear. For example, we might expect task and domain specificity, and differences due to the quality and the proportion of categorical features in a dataset \citep{Valdez2021}. To the best of our knowledge, none of the papers, exploring the encoding techniques have controlled for the characteristics of the categorical variables, their number or their importance for models.

Throughout the paper, we follow Pargent and colleagues \citep{Pargent2022} in grouping the encoding techniques into target-agnostic and target-based. Target-agnostic encoding schemes replace category levels with numbers that are unrelated to the target or any other feature in a dataset. These constitute the traditional encoding schemes, transforming feature levels into random integers (label/ordinal/integer encoding) or with a vector of numbers, typically of size $N$ or $N-1$ (indicator encoding schemes, including one-hot encoding, Helmert contrast coding, sum encoding and others), where $N$ is the cardinality of a feature – the number of its unique values. 

One disadvantage of the integer/label encoding is that it introduces an ordinal relationship between levels, which is probably absent in the data. This might be particularly damaging for linear or deep learning models, especially with high cardinality features. In contrast, the major disadvantage of indicator encoders like one-hot or Helmert contrast coding is that they increase the number of columns in a dataset. For a task with a relatively small number of observations and a large proportion of categorical features, particularly those with high cardinality, this leads to the ‘curse of dimensionality’ problem. With large datasets, we will probably have sufficient samples to overcome this issue; however, processing wide datasets significantly increases training duration and memory capacity demands.

More recently, Micci-Barreca has proposed the first target-based encoding scheme to address the shortcomings of the traditional target-agnostic approaches \citep{Barreca2001}. As follows from its name, target-based encoder replaces each level of a category with a number based on the corresponding target values. For binary problems, the sizes of the original and the encoded datasets remain the same. For multiclass tasks, however, each categorical variable is transformed into a feature vector of size $M-1,$ where $M$ is the number of target classes. Therefore, multiclass datasets with high cardinality target may potentially lead to dimensionality issues similar to those described for target-agnostic indicator encoders. Other two interrelated problems of target-based encoders reportedly include data leakage and the general tendency to overfit. General overfitting occurs when the training data and the inference data follow different distributions – since the encoding algorithm is fitted only on the former. This is known as domain adaptation, occurring due to target shift and conditional (non-target variable) shift \citep{Zhang2013}. Data leakage with target-based encoders is a more extreme case of overfitting that takes place when the encoded categorical feature has almost all unique values \citep{Prokhorenkova2017}. The encoder creates a strong correlation between such feature and the target, since each level is replaced with a statistic calculated from the target. However, regularization techniques have been shown to successfully overcome these issues \citep{Mougan2022}, \citep{Pargent2022}.

To summarize, target-agnostic and target-based encoding schemes have conceptually distinct approaches to transforming categorical variables. Limited empirical evidence fails to recommend one approach over the other. Therefore, the patterns present in real datasets are still poorly understood, and expectations that, for example, target-based encoding would improve model performance are not grounded on any solid evidence and requires further study. In this paper, we explored the effects of encoding, using 5 encoding techniques: target-agnostic integer encoding (label encoding) used as our baseline, two types of target-agnostic indicator encoding (one-hot encoding and Helmert contrast coding) and two types of target-based encoding (target and weight of evidence encoding). We expected to observe similar encoding effects within each group: for example, if target encoder had a significant effect on the outcome, we were likely to observe similar patterns for weight of evidence encoder (WoE). Having two conceptually similar encoders allowed exploring in detail generalizability of any observed effects. Below we discuss each encoding algorithm used in the study in greater detail.

\subsection{Encoding algorithms}

\textbf{Label encoder} transforms feature categories (levels) into integers, ranging from 1 to $N$, where $N$ is a feature cardinality. This type of encoding is also known as ordinal encoding because it introduces an ordinal relationship between feature levels. If a true order is present in the data, such scheme might be beneficial as it preserves more information for fitting a model. Although most encoders permit specifying the order manually or including it as meta-information (for example, passing a mapping dictionary in the scikit-learn category encoders module \citep{McGinnis2018}), this task is impractical or not feasible in most cases. First, few feature categories have true ordinal scale; second, it requires expertise and close familiarity with the data to define appropriate scales for features with string values. For this experiment, we relied on a pseudorandom initiation of the levels.

\textbf{One-hot encoder }(OHE) is a target-agnostic indicator encoder that replaces categorical features with sparse vectors, containing all zeros except for a single 1 in $i^{th}$ place of the correct level, as in $x_i = \underbrace{[0, \dots, 1^i, \dots, 0]}_N$. The size of the vector is typically $N$ or $N-1$ (dropping the first category), where $N$ is the feature cardinality. In our experiments, we used the default OHE setting in Category Encoders module, producing $N$ columns for each feature of cardinality $N$.

\textbf{Helmert contrastive coding} is a target-agnostic indicator technique that transforms a categorical feature into $N$ new features ($N$ is this feature cardinality). It is often used in cases when we expect some type of monotony in the response variable \citep{Carey2013}. For example, when we treat patients with different dosages of the same medicine to identify the minimal effective dose \citep{Berg2022}, or when we compare stimuli effects at different time points after their presentation \citep{Carey2013}. Helmert contrast coding assumes an underlying meaning in the categories’ order, transforming their values using a type of one-vs-rest method, where ‘rest’ stands for ‘all subsequent levels’. Specifically, it replaces categories with numerical values from a contrastive matrix, so that a regression model would compare each level of a categorical variable to the mean of the subsequent levels, thus identifying a turning point in a sequence.

\textbf{Target encoder} 
uses target statistics to transform feature categories. Specific strategy depends on the type of problem. For regression tasks, each category is replaced with the conditional target mean \citep{Pargent2022} – the mean of the target values for all instances of that category. For classification tasks, each level is replaced with the conditional posterior probability, or the conditional relative frequency of the target. Micci-Barreca \citep{Barreca2001} blends the conditional target mean with the target mean across all categories; similarly, the conditional posterior probability of the target for a category is blended with the prior probability of the target across all categories:

\begin{equation*}
 x_i = \lambda(n_i)\frac{n_{iY}}{n_i} + (1-\lambda(n_i))\frac{n_Y}{n}, 
\end{equation*}
 where $n_{iY}$ is the target average for samples within category $i$, $n_i$ is the number of training examples in category $i$, $n_Y$ is the target average, $n$ -- the 
number of all training examples, and $\lambda(n_i)$ -- the weighting factor. 
The latter variable $\lambda(n_i)$ is calculated as follows:

\begin{equation*}
 \lambda(n_i) = \frac{1}{1 + e^{-\frac{n_i-min\_samples\_leaf}{smoothing}}},
\end{equation*}
 where $min\_samples\_leaf$ and $smoothing$ are tunable parameters.

The category encoders module requires a PolynomialWrapper to apply target encoding to multiclass datasets: Each categorical feature is transformed into a vector of size $M$, where $M$ is the number of target classes. The probabilities for each feature level are calculated, following the classical one-vs-rest scenario for each target class. In our experiments, we used the default regularization settings: $f$ = 1.0 and $k$ = 1.

\textbf{Weight of evidence encoding} (WOE)
is a target-based encoding technique that replaces feature levels based on the amount of empirical support this category gives for the probability of one target class over another \citep{Zeng2014}. The concept comes from Bayesian statistics and is computed by taking a  natural logarithm of a Bayesian likelihood ratio \citep{Good85}:

\begin{equation*}
WOE_i = \ln{\frac{proportion\_positive_i}{proportion\_negative_i}}, 
\end{equation*}
 where $proportion\_positive_i = \frac{n_i^+}{n^+}$ is the proportion of positive examples in the target across category $i$ samples,
$proportion\_negative_i = \frac{n_i^-}{n^-}$  is the proportion of negative examples in the target across category $i$ samples,
$n_i^{-(+)}$ --  the number of negative (positive) examples with the category $i$,
 and $n^{-(+)}$ is the total number of negative (positive) examples.

WOE was designed for binary problems, requiring adaptations for regression tasks (via discretization) and for multiclass tasks (by creating a feature vector with multiple one-vs-rest comparisons). With binary problems, for each feature level we first compute the number of target events and non-events (or the number of instances with target class 0 vs. 1; for example, for categories ‘red’ and ‘blue’ in ‘color’ feature, we have 60 vs. 30 cases where the target is 0 as opposed to 1). We calculate the proportion of these two values in the total number of non-events vs. events (for example, we had the total of 1000 samples, with 780 vs. 220 samples of class labels 0 and 1, respectively; 
 $\frac{60}{780} \approx 0.077$ vs. $\frac{30}{220} \approx 0.136$). After that we compute the Bayesian likelihood ratio and take its natural logarithm (in our case, $ \ln{\frac{0.077}{0.136}} \approx - 0.569$, which replaces the category ‘red’ in the ‘color’ feature). As with target encoder, we used PolynomialWrapper with multiclass problems, and the default regularization settings.

\subsection{Present study}
The main aim of this study is to explore the effects of encoding in a large pull of real world classification problems. We used an exhaustive sample of suitable binary and multiclass OpenML datasets for our experiments. We fitted H2O AutoML to each dataset, extracting model performance metrics, training/inference duration info and feature importances for analysis. These data we supplemented with the features engineered from OpenML metadata (see details in Section 2 Method). We fitted several LMMs, with model performance metrics and training duration as targets, task ID as a random effect and other features as fixed effects. Because we predicted possible differences, depending on the model type, we compared three different architectures from those fitted by H2O AutoML: a linear model, a non-linear tree-based model (homogeneous ensemble) and a stacked (heterogeneous) ensemble, comprising the first two estimators (along with other models) as base learners. We were also interested in model training/inference efficiency. Therefore, we investigated, how the training duration changed as a function of encoding type.

The collected experimental data allowed addressing the following key research questions:

\textbf{1. Does the encoding type influence the predictive power of a heterogeneous ensemble?}

We predict that the encoding type may influence some base estimators. However, if small, we expect these effects to become negligible in a stacked ensemble.

\textbf{2. Does the type of encoding influence the predictive power of an individual base estimator? Can we observe linear vs. non-linear effects?}

We hypothesize that target-based encoding will increase the predictive power of base models, given the recent evidence of their successful performance in some real world tasks \citep{Bourdonnaye2021}, \citep{Hien2020}, \citep{Potdar2017}, \citep{Seca2021}, \citep{Valdez2021}.

We also expect differences between linear and non-linear models. Specifically, we predict a negative effect of label encoding on linear models, because this encoding scheme introduces an ordinal relationship between feature levels, which is likely not present in datasets; we expect these effects to be particularly prominent in high cardinality datasets. In contrast, for tree-based models we expect to see no difference between label encoder and other target-agnostic encoders.

\textbf{3. Does the type of encoding influence training efficiency?}

We predict that label encoding will likely lead to the most time-efficient training, because this scheme neither expands the feature space, nor involves computing statistics. In contrast, OHE and Helmert contrast coding will likely be significantly less efficient than any other encoding scheme, particularly in binary tasks: model training efficiency suffers from the increased feature space.

In multiclass problems, however, we expect little difference between OHE/Helmert contrast coding and target-based encoders: the latter increase the size of each categorical feature proportionally to the number of target classes.

\section{Method}
\begin{figure*}[]
\centering
\includegraphics[width=0.8\textwidth]{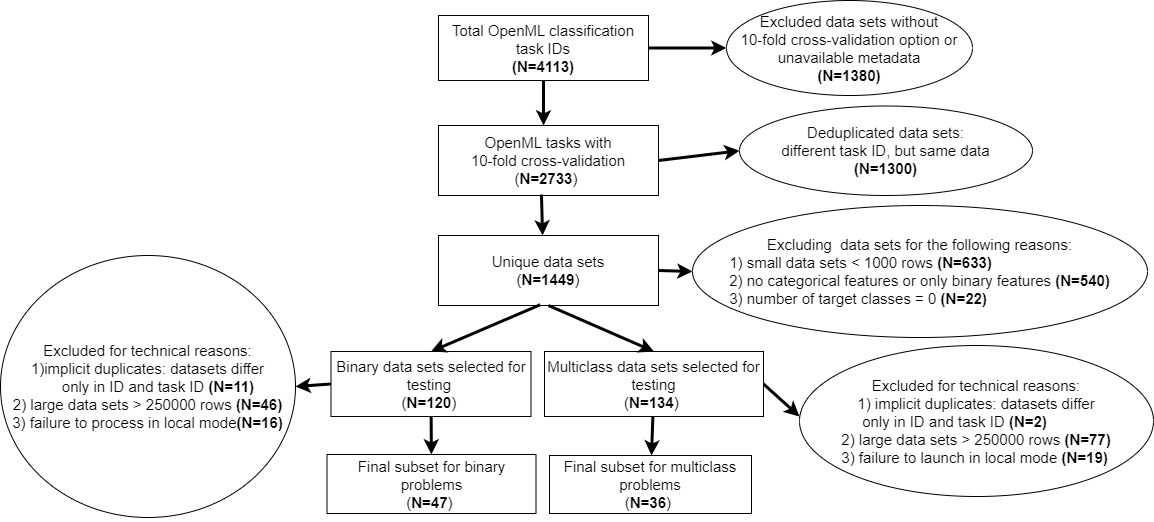}
\caption{Dataset selection flowchart.}
\label{fig1}
\end{figure*}
We investigated the effects of encoding on model predictive power and on training/inference durations. In order to control for possible confounding effects of biased data selection, we exhaustively sampled binary and multiclass classification problems from OpenML \citep{Vanschoren2014}. With each suitable dataset, we ran H2O AutoML \citep{hdo}, collecting model evaluation metrics, feature importance information and duration estimates. These data were coupled with each dataset’s metadata information, describing the various characteristics of categorical features, including their proportion and the number of categories in a feature. We fitted a series of LMMs to the resulting data, in order to explore the various effects of encoding. Subsections below describe data preparation in greater detail.

\subsection{Data}

For data sources, we used OpenML – an open platform for sharing datasets and running collaborative research in ML, focusing particularly on tabular data. An individual dataset may have different associated tasks: OpenML reports the total of 4256 datasets and 260755 associated tasks. In the majority of cases, the original data remained the same, but the problem was formulated differently. When different tasks referred to the same version of a dataset, we kept a single copy of the data. Alternatively, the authors may have transformed the data in some way: for example, changed a multiclass target into a binary target, created a version of the dataset with only a subset of original features or samples available for analysis. These transformations are treated as independent datasets in OpenML. For the purposes of our research, we followed the same approach, sampling data in both their raw and filtered forms. 

We specified ‘supervised classification’ task type in the search form of the platform API, resulting in a subset of 4113 OpenML unique binary and multiclass problems. We then extracted metadata associated with task IDs and performed additional filtering. This included dataset deduplication (identical data, but different task IDs), removal of data without 10-fold cross-validation option, etc. The flowchart in Figure 1 describes the entire selection procedure.

%\begin{figure}[h]
%\centering
%\includegraphics[width=0.8]{fig1}
%\caption{Data set selection flowchart.}
%\end{figure} Dataset selection flowchart.

% figure 1

%\begin{figure}[tb]
%\centering
%\includegraphics[width=0.4\textwidth]{1_}
%\caption{Dataset selection flowchart.}
%\label{fig1}
%\end{figure}

The resulting sample of unique datasets included 254 entries: 120 binary and 134 multiclass problems. During testing, however, this subset was additionally filtered due to technical issues, such as failure to download data from the OpenML repository with the specified task IDs, or insufficient CPU resources to process dataset in local mode due to its large size. The final selection, for which we could collect experimental data, included 47 binary datasets and 36 multiclass datasets.
\subsection{Experimental setup}

This section describes all the stages of data pre-processing and modelling that we followed to address the research questions.

\subsubsection{Data pre-processing}
For all tasks we followed the same preprocessing pipeline. First, a dataset was downloaded, using OpenML AutoML Benchmark framework \citep{Gijsbers2019}. Second, we imputed missing numerical values with a column mean, and missing categorical values with a ‘new category’ label. Finally, we created multiple versions of the same dataset, encoding all categorical variables 5 ways, using the category encoders module from the scikit-learn-contrib package \citep{McGinnis2018}: label (ordinal) encoder, OHE, Helmert coding, target encoder, WOE. With all encoders we used the default settings.

\subsubsection{Models}
We fitted H2O AutoML to all selected binary and multiclass tasks. The advantage of using a high-performing end-to-end solution is that it allows testing multiple hypotheses within a single framework, controlling for possible confounds. H2O AutoML fits individual linear and non-linear models toped up with stacked ensembles\citep{Wolpert92}, orchestrating individual estimators as base learners. The framework relies on a specific type of stacked ensemble – super learner\citep{Laan2007} – which uses cross-validation for creating a metadataset from base model predictions. During experiments, we logged information about training and inference durations, predictive performance on unseen data and estimates of feature importances across different types of models.

We specified the types of base models from a selection of algorithms available in H2O AutoML, and set the maximum number of base estimators to 5 models. These included H2O distributed implementations of generalized linear model with regularization (GLM), extreme gradient boosting (XGBoost), gradient boosting machine (GBM), distributed random forest (DRF, comprising distributed version of random forest and the H2O implementation of extremely randomized trees). H2O AutoML predefines model hyperparameters and training order. In all cases, the system fitted two XGBoost models with different hyperparameters, and a single instance of other algorithms.

Enabling stacked ensembles option in H2O AutoML, results in two super learner ensembles: ‘best of family’ (a stacked ensemble that is trained on predictions from a single best-performing model of the same type) and ‘all’ (stacking predictions from all trained models). Because we only had 5 base models, both ensembles were very close in terms of their composition and predictive power. For the purposes of our study, we report performance of the ensemble with all 5 trained models used as base learners. Experimental data were collected on a desktop computer Inter Core i7-8700 CPU@ 3.2GHz x 12 32 GiB. Maximal training duration was set to 1 hour per cross-validation fold.
\subsubsection{Metrics and measurements}
In the study, we explored the effects of encoding in two types of classification problems – in binary and in multiclass tasks – requiring two different target metrics. For binary tasks, we reported the area under the receiver operating characteristic curve (AUC-ROC), and for multiclass tasks – the log-loss estimate.
Importantly, the absolute values of these metrics have the opposite interpretation. AUC-ROC estimates probabilities, and its values are in the [0; 1] range, where higher values indicate better model performance. Log-loss measures the divergence of the prediction from the true value, hence, its values are in the  $[0, + \infty )$ interval, where smaller values indicate better model performance. 

Additionally, we aggregated information on model training and inference durations, to estimate the training cost and its scaling, particularly with encoding schemes, resulting in a great expansion of the original feature space, such as OHE in case of high cardinality.

Finally, we logged information about feature importances across models, reflecting the influence of categorical variables on specific models. We predicted an interaction between the encoding schemes and feature importance estimates: encoding type was more likely to influence the outcome when the encoded features were important for the models. We, thus, extracted feature importance information from one linear and one non-linear model. For the GLM, feature importances come from the estimates of the model standardized coefficients. The XGBoost estimates of variable importances are based on the gains of their respective loss functions during tree construction. For analysis, we created the following two types of feature importance estimates: 1) the total number of non-binary categorical features in the top 5 most important features for a model (integer in a [0; 5] interval); 2) summed scaled importance of all non-binary categorical features (in a [0; 1] interval).

\subsubsection{Metadata}
We created three variables that characterize prevalence of categorical information in a dataset, using metadata information from OpenML: \emph{proportion\_cat} – proportion of categorical features in a dataset, \emph{avg\_cardinality} – the average number of levels across all non-binary categorical features in a dataset, \emph{dev\_cardinality} – standard deviation around the mean for the same cardinality estimate.

\subsubsection{Experimental data}
Dataset that we used for fitting LMMs integrated information, coming from three sources: 1) H2O AutoML outputs (metrics of the stacked ensemble, GLM and XGBoost; training duration values; feature importances for GLM and XGBoost); 2) metadata from OpenML (task ID and number of target classes for multiclass problems; plus \emph{proportion\_cat}, \emph{avg\_cardinality} and \emph{dev\_cardinality}); 3) additional \emph{high\_cardinality} feature engineered during visual inspection of the data (see Figure 2).

% figure 2
\begin{figure}[h]
\centering
\includegraphics[width=0.5\textwidth]{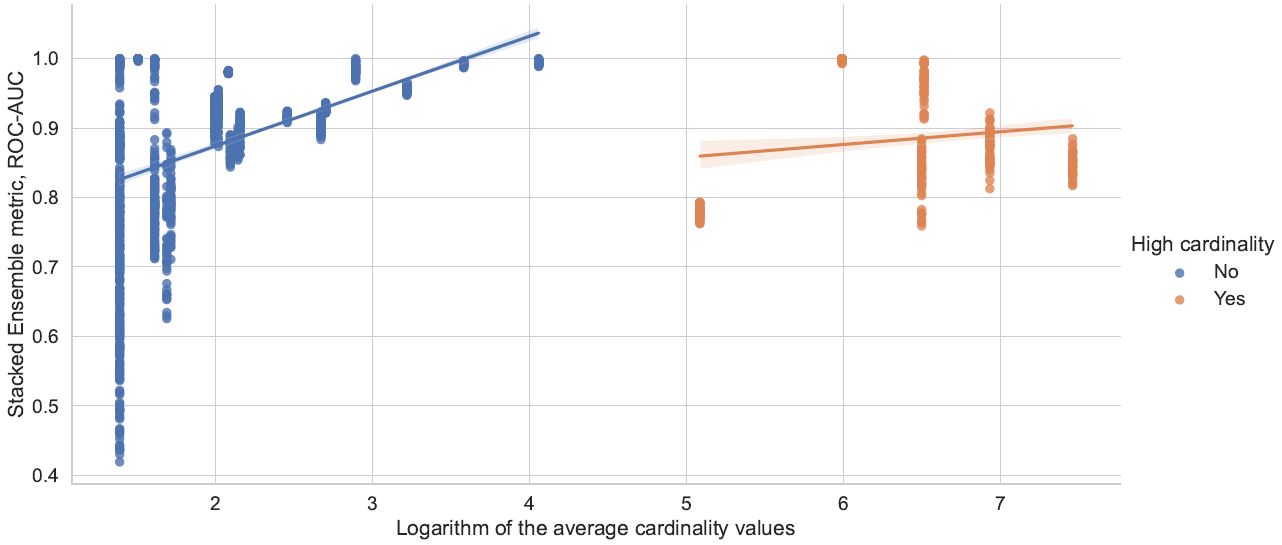}
\caption{Stacked ensemble metric as a function of the datasets’ average cardinality values, grouped into low- and high-cardinality datasets.}
\end{figure}
\section{Results}
We fitted multiple LMMs to the experimental data, addressing the research questions discussed in section \emph{1 Introduction}. The subsections below follow the same order as did these research questions. We analyzed binary and on multiclass problems separately, because we predicted different patterns – at least for some models.

For both multiclass and binary datasets, we fitted LMMs, using each of the following 4 variables as a target: 1) metric of a stacked ensemble; 2) metric of an individual linear model (GLM); 3) metric of an individual non-linear model (XGBoost); 4) training duration. We treated the various characteristics of a dataset and the type of encoding as fixed effects, and assumed random intercept and slope (and their correlation) for each dataset across the encoding types. For multiclass problems, we additionally included information about the number of target classes. With each target we fitted two types of models – with and without two-way interaction terms. Following backward elimination procedure, we excluded non-significant predictors and predictor interactions. The final most parsimonious model was selected based on Akaike information criterion (AIC). For all types of analysis we used pymer4 package \citep{Jolly2018}, allowing to run R-like \citep{R2014} style of coding LMM formulas within the Python ecosystem.

\subsection{\emph{Does the encoding type influence the predictive power of a heterogeneous ensemble, such as a super learner stacked ensemble?}}
\noindent
\emph{Binary datasets}

We fitted LMMs with the stacked ensemble metric (\emph{SE\_metric}) as a dependent variable. The best model included only individual effects without interaction terms (see output in Table 1).

% table 1
\begin{table}[h]
\fontsize{10}{15}\selectfont 
\resizebox{\linewidth}{!}{
\begin{tabular}{|cc|r|r|r|c|}
\hline
\multicolumn{2}{|c|}{Fixed effects} & \multicolumn{1}{l|}{Estimate} & \multicolumn{1}{l|}{95\% Confidence interval} & \multicolumn{1}{l|}{\emph{p}-value} & \multicolumn{1}{l|}{Significance level} \\ \hline
\multicolumn{2}{|c|}{(Intercept)} & 0.835 & {[}0.735; 0.936{]} & \textless{}0.001 & *** \\ \hline
\multicolumn{1}{|c|}{\multirow{4}{*}{\begin{tabular}[c]{@{}c@{}}\emph{Encoder}: \\    target encoder vs.\end{tabular}}} & Helmert & 0.002 & {[}-0.003; 0.006{]} & 0.451 &  \\ \cline{2-6} 
\multicolumn{1}{|c|}{} & OHE & 0.004 & {[}-0.000; 0.008{]} & 0.080 & . \\ \cline{2-6} 
\multicolumn{1}{|c|}{} & WOE & 0.001 & {[}-0.003; 0.005{]} & 0.641 &  \\ \cline{2-6} 
\multicolumn{1}{|c|}{} & label & 0.002 & {[}-0.006; 0.002{]} & 0.295 &  \\ \hline
\multicolumn{2}{|c|}{\emph{High cardinality} (binary)} & -0.333 & {[}-0.528; -0.139{]} & 0.002 & ** \\ \hline
\multicolumn{2}{|c|}{\emph{Proportion of categorical  features} (continuous)} & -0.204 & {[}-0.294; -0.115{]} & \textless{}0.001 & *** \\ \hline
\multicolumn{2}{|c|}{\emph{Average cardinality} (continuous)} & 0.071 & {[}0.031; 0.112{]} & 0.001 & ** \\ \hline
\end{tabular}}
\caption{\label{tab:somelabel}Predicting variability in the stacked ensemble metric for binary tasks.} 
\end{table}
The most significant predictor of the \emph{SE\_metric} was the proportion of categorical features in a dataset (variable \emph{proportion\_cat}; $p$-value $<$ 0.001, estimate = -0.204; 95\% CI [-0.294; -0.115]). This variable had a strong negative effect on the outcome, suggesting that overall problems with a greater number of categorical features might be more challenging compared to tasks with a higher proportion of numerical features. Dataset average cardinality (variable \emph{avg\_cardinality}; $p$-value = 0.001, estimate = 0.071; 95\% CI [0.031; 0.112]) had a positive effect on the predictive power of the ensemble, possibly indicating that a variable with greater number of levels contains more information, which is beneficial for accurate classification. In contrast, \emph{high\_cardinality} had a negative effect on the model predictive power ($p$-value = 0.002, estimate = -0.333; 95\% CI [-0.528; -0.139]). This suggests that above a certain threshold – in our sample, when categorical features have, on average, over 90 categories – additional levels contribute to noise, decreasing model performance. The encoder type did not have a significant effect on the outcome, although we observed a trend to significance for OHE, compared to the baseline ($p$-value = 0.08, estimate = 0.004; 95\% CI [-0.000; 0.008]). However, the diagnostic Q-Q plots revealed heavy tails (that we could not resolve by data transformations), probably because of the negative skewness in the response variable, i.e., the median of the \emph{SE\_metric} is around 0.90. Thus, we need to be cautious when interpreting marginally significant effects.

\noindent
\emph{Multiclass datasets}
\newline
\indent The ensemble log-loss metric was predicted by the type of encoding, the characteristics of categorical features in a dataset, and the number of classes in the target variable. We followed backward elimination procedure, resulting in a LMM with only two predictors: type of encoding and the number of target classes (see Table 2). 
% table 2
\begin{table}[h]
\fontsize{10}{15}\selectfont 
\resizebox{\linewidth}{!}{
\begin{tabular}{|cc|r|r|r|c|}
\hline
\multicolumn{2}{|c|}{Fixed effects} & \multicolumn{1}{c|}{Estimate} & \multicolumn{1}{c|}{95\% Confidence interval} & \multicolumn{1}{c|}{\emph{p}-value} & Significance level \\ \hline
\multicolumn{2}{|c|}{(Intercept)} & -0.777 & {[}-1.425; -0.128{]} & 0.025 & * \\ \hline
\multicolumn{1}{|c|}{\multirow{4}{*}{\begin{tabular}[c]{@{}c@{}}\emph{Encoder}: \\    label encoder vs.\end{tabular}}} & target & 0.006 & {[}-0.021; 0.033{]} & 0.684 &  \\ \cline{2-6} 
\multicolumn{1}{|c|}{} & WOE & -0.002 & {[}-0.023; 0.019{]} & 0.834 &  \\ \cline{2-6} 
\multicolumn{1}{|c|}{} & OHE & -0.012 & {[}-0.020; -0.003{]} & 0.015 & * \\ \cline{2-6} 
\multicolumn{1}{|c|}{} & Helmert & -0.010 & {[}-0.018; -0.003{]} & 0.011 & * \\ \hline
\multicolumn{2}{|c|}{\emph{Number of target classes}} & 0.750 & {[}0.432; 1.068{]} & \textless{}0.001 & *** \\ \hline
\end{tabular}}
\caption{Predicting variability in the stacked ensemble metric for multiclass tasks.} 
\end{table}

The number of target classes was the key factor responsible for variability in the \emph{SE\_metric} values ($p$-value $<$ 0.001, estimate = 0.75; 95\% CI [0.432; 1.068]): predictably, problems with a greater number of classes were more challenging even for heterogeneous algorithms. Encoding type had a small, but significant negative effect on the loss metric for two types of encoding: OHE ($p$-value = 0.015, estimate = -0.012; 95\% CI [-0.020; -0.003]) and Helmert ($p$-value = 0.011, estimate = -0.010; 95\% CI [-0.018; -0.003]). This suggests that both encoders improve ensemble performance (with logloss the less is the better) compared to target-based encoders and label encoder.

\subsection{\emph{Does the type of encoding influence the predictive power of an individual base estimator? Can we observe linear vs. non-linear effects?}}

We will first focus on the linear estimator performance (GLM) and then summarize the results for the non-linear estimator (XGBoost). In both cases, we will discuss binary and multiclass problems separately.

\subsubsection{Linear estimator: GLM performance}
\noindent
\emph{Binary datasets}

We used a very similar model fit procedure as above, with the following minor modifications: \emph{GLM\_metric }was our target; for categorical feature importances we used only estimates from the linear model, excluding those from the XGBoost model. The most parsimonious model included only individual predictors without two-way interactions (see Table 3).

% table 3

\begin{table}[h]
\fontsize{10}{15}\selectfont 
\resizebox{\linewidth}{!}{
\begin{tabular}{|cc|r|r|r|c|}
\hline
\multicolumn{2}{|c|}{Fixed effects} & \multicolumn{1}{c|}{Estimate} & \multicolumn{1}{c|}{95\% Confidence interval} & \multicolumn{1}{c|}{\emph{p}-value} & Significance level \\ \hline
\multicolumn{2}{|c|}{(Intercept)} & 0.953 & {[}0.879; 1.027{]} & \textless{}0.001 & *** \\ \hline
\multicolumn{1}{|c|}{\multirow{4}{*}{\begin{tabular}[c]{@{}c@{}}\emph{Encoder}: \\    target encoder vs.\end{tabular}}} & Helmert & -0.001 & {[}-0.006; 0.004{]} & 0.699 &  \\ \cline{2-6} 
\multicolumn{1}{|c|}{} & OHE & 0.000 & {[}-0.005; 0.005{]} & 0.955 &  \\ \cline{2-6} 
\multicolumn{1}{|c|}{} & WOE & -0.000 & {[}-0.005; 0.004{]} & 0.889 &  \\ \cline{2-6} 
\multicolumn{1}{|c|}{} & label & -0.019 & {[}-0.034; -0.004{]} & 0.015 & * \\ \hline
\multicolumn{2}{|c|}{\emph{High cardinality} (binary)} & -0.236 & {[}-0.369; -0.103{]} & 0.001 & ** \\ \hline
\multicolumn{2}{|c|}{\emph{Proportion of categorical  features} (continuous)} & -0.341 & {[}-0.450; -0.232{]} & \textless{}0.001 & *** \\ \hline
\multicolumn{2}{|c|}{\emph{Deviation of cardinality around the mean} (cont.)} & 0.038 & {[}0.018; 0.059{]} & 0.001 & *** \\ \hline
\end{tabular}}
\caption{Predicting variability in the GLM metric for binary tasks.} 
\end{table}

The most significant predictor of the \emph{GLM\_metric} was \emph{proportion\_cat} ($p$-value $<$ 0.001, estimate = -0.341; 95\% CI [-0.450; -0.232]), indicating that tasks with a higher proportion of categorical features are more challenging for the linear estimator. The second most important predictor was \emph{dev\_cardinality} – the spread of variability in the number of levels across categorical features. Similarly to the \emph{avg\_cardinality} counterpart for the stacked ensemble model, it demonstrated a positive effect on the target ($p$-value = 0.001, estimate = 0.038; 95\% CI [0.018; 0.059]). Since \emph{dev\_cardinality} values increase linearly with the average cardinality of the dataset, we assume similar interpretation of the positive effect of this variable on the target: a greater number of levels in categorical features contains more information useful for the model. This effect, however, is leveraged by the \emph{high\_cardinality} predictor ($p$-value = 0.001, estimate = -0.236; 95\% CI [-0.369; -0.103]): all things being equal, datasets with high average cardinality had a lower \emph{GLM\_metric}. Finally, label encoding had a small yet statistically significant negative effect on the outcome ($p$-value = 0.015; estimate = -0.019; 95\% CI [-0.034; -0.004]). This confirms our prediction that label encoder in its default setting might decrease the predictive power of linear models by introducing a non-existent ordinal relationship between feature levels. \newline 

\noindent \emph{Multiclass datasets}
\newline
\indent The most parsimonious model included the same main effects as the model summarized in Table 2. The number of target classes had the most significant effect on the \emph{GLM\_metric} ($p$-value $<$ 0.001, estimate = 0.985; 95\% CI [0.700; 1.270]): the more classes the model predicts, the more challenging the task (see Table 4). 

% table 4
\begin{table}[h]
\fontsize{10}{15}\selectfont 
\resizebox{\linewidth}{!}{
\begin{tabular}{|cc|r|r|r|c|}
\hline
\multicolumn{2}{|c|}{Fixed effects} & \multicolumn{1}{c|}{Estimate} & \multicolumn{1}{c|}{95\% Confidence interval} & \multicolumn{1}{c|}{\emph{p}-value} & Significance level \\ \hline
\multicolumn{2}{|c|}{(Intercept)} & -1.035 & {[}-1.618; -0.452{]} & 0.001 & ** \\ \hline
\multicolumn{1}{|c|}{\multirow{4}{*}{\begin{tabular}[c]{@{}c@{}}\emph{Encode}r: \\    label encoder vs.\end{tabular}}} & target & -0.112 & {[}-0.191; -0.033{]} & 0.009 & ** \\ \cline{2-6} 
\multicolumn{1}{|c|}{} & WOE & -0.112 & {[}-0.190; -0.033{]} & 0.008 & ** \\ \cline{2-6} 
\multicolumn{1}{|c|}{} & OHE & -0.107 & {[}-0.188; -0.027{]} & 0.013 & * \\ \cline{2-6} 
\multicolumn{1}{|c|}{} & Helmert & -0.049 & {[}-0.155; 0.057{]} & 0.368 &  \\ \hline
\multicolumn{2}{|c|}{\emph{Number of target classes}} & 0.985 & {[}0.700; 1.270{]} & \textless{}0.001 & *** \\ \hline
\end{tabular}}
\caption{Predicting variability in the GLM metric for multiclass tasks.} 
\end{table}
We additionally observed a significant effect of the encoding type. Both target-based encodings and OHE had a similar positive effect on the outcome (target encoding: $p$-value = 0.009, estimate = -0.112, 95\% CI [-0.191; -0.033]; WOE: $p$-value=0.008; estimate = -0.112, 95\% CI [-0.190; -0.033]; OHE: $p$-value = 0.013, estimate = -0.107; 95\% CI [-0.188; -0.027]) compared to label encoding. 

\subsubsection{Non-linear model: XGBoost performance}
\noindent
\emph{Binary datasets}

We fitted GLM, using XGBoost model metric (\emph{XGBoost\_metric}) as our target, preserving most of our original setup for the independent predictors, but using only estimates from the non-linear model for the categorical feature importances, excluding those from the GLM model. The most parsimonious model was a LMM with only individual variables without two-way interaction terms (see Table 5).
% table 5
\begin{table}[h]
\fontsize{10}{15}\selectfont 
\resizebox{\linewidth}{!}{
\begin{tabular}{|cc|r|r|r|c|}
\hline
\multicolumn{2}{|c|}{Fixed effects} & \multicolumn{1}{c|}{Estimate} & \multicolumn{1}{c|}{95\% Confidence interval} & \multicolumn{1}{c|}{\emph{p}-value} & Significance level \\ \hline
\multicolumn{2}{|c|}{(Intercept)} & 0.831 & {[}0.724; 0.938{]} & \textless{}0.001 & *** \\ \hline
\multicolumn{1}{|c|}{\multirow{4}{*}{\begin{tabular}[c]{@{}c@{}}\emph{Encoder}: \\    target encoder vs.\end{tabular}}} & Helmert & 0.001 & {[}-0.004; 0.007{]} & 0.620 &  \\ \cline{2-6} 
\multicolumn{1}{|c|}{} & OHE & -0.000 & {[}-0.007; 0.007{]} & 0.993 &  \\ \cline{2-6} 
\multicolumn{1}{|c|}{} & WOE & 0.001 & {[}-0.003; 0.006{]} & 0.595 &  \\ \cline{2-6} 
\multicolumn{1}{|c|}{} & label & -0.001 & {[}-0.006; 0.003{]} & 0.548 &  \\ \hline
\multicolumn{2}{|c|}{\emph{High cardinality} (binary)} & -0.345 & {[}-0.553; -0.138{]} & 0.003 & ** \\ \hline
\multicolumn{2}{|c|}{\emph{Proportion of categorical   features} (continuous)} & -0.195 & {[}-0.291; -0.100{]} & \textless{}0.001 & *** \\ \hline
\multicolumn{2}{|c|}{\emph{Average cardinality} (continuous)} & 0.068 & {[}0.025; 0.111{]} & 0.004 & ** \\ \hline
\end{tabular}}
\caption{Predicting variability in the XGBoost metric for binary tasks.} 
\end{table}

As with other binary tasks, the \emph{proportion\_cat} feature had the most significant impact on the \emph{XGBoost\_metric}, demonstrating a negative relationship between the variables ($p$-value $<$ 0.001, estimate = -0.195; 95\% CI [-0.291; -0.100]). The \emph{avg\_cardinality} had a positive effect ($p$-value = 0.004, estimate = 0.068; 95\% CI [0.025; 0.111]) on the target, but the effect was mitigated by the negative relationship with \emph{high\_cardinality} ($p$-value = 0.003, estimate = -0.345; 95\% CI [-0.553; -0.138]). We offer the same interpretations of these effects as for the stacked ensemble and GLM models above.\newline

\noindent
\emph{Multiclass datasets}
\newline
\indent As with binary datasets, the LMM model for multiclass problems did not reveal any effects of encoding on the outcome variable \emph{XGBoost\_metric} (see Table 6). 
% table 6
\begin{table}[h]
\fontsize{10}{15}\selectfont 
\resizebox{\linewidth}{!}{
\begin{tabular}{|cc|r|r|r|c|}
\hline
\multicolumn{2}{|c|}{Fixed effects} & \multicolumn{1}{c|}{Estimate} & \multicolumn{1}{c|}{95\% Confidence interval} & \multicolumn{1}{c|}{\emph{p}-value} & Significance level \\ \hline
\multicolumn{2}{|c|}{(Intercept)} & 0.193 & {[}-0.690; 1.077{]} & 0.671 &  \\ \hline
\multicolumn{1}{|c|}{\multirow{4}{*}{\begin{tabular}[c]{@{}c@{}}\emph{Encoder}: \\    label encoder vs.\end{tabular}}} & target & 0.008 & {[}-0.028; 0.044{]} & 0.667 &  \\ \cline{2-6} 
\multicolumn{1}{|c|}{} & WOE & -0.007 & {[}-0.031; 0.018{]} & 0.597 &  \\ \cline{2-6} 
\multicolumn{1}{|c|}{} & OHE & -0.004 & {[}-0.026; 0.017{]} & 0.691 &  \\ \cline{2-6} 
\multicolumn{1}{|c|}{} & Helmert & -0.005 & {[}-0.015; 0.006{]} & 0.402 &  \\ \hline
\multicolumn{2}{|c|}{\emph{High cardinality} (binary)} & 1.802 & {[}0.513; 3.092{]} & 0.010 & * \\ \hline
\multicolumn{2}{|c|}{\emph{Proportion of categorical features} (continuous)} & -0.758 & {[}-1.340; -0.175{]} & 0.016 & * \\ \hline
\multicolumn{2}{|c|}{\emph{Average cardinalit}y (continuous)} & -0.511 & {[}-0.859; -0.164{]} & 0.007 & ** \\ \hline
\multicolumn{2}{|c|}{\emph{Number of target classes} (integer)} & 0.942 & {[}0.592; 1.293{]} & \textless{}0.001 & *** \\ \hline
\end{tabular}}
\caption{Predicting variability in the XGBoost metric for multiclass tasks.} 
\end{table}

The most parsimonious model included only main effects. The number of target classes had the most significant detrimental effect on the target ($p$-value $<$ 0.001, estimate = 0.942; 95\% CI [0.592; 1.293]). \emph{High\_cardinality} datasets were overall significantly more problematic for the XGBoost algorithm ($p$-value = 0.01, estimate = 1.802; 95\% CI [0.513; 3.092]). In contrast, both the proportion of categorical features and the average cardinality of the dataset had a positive effect on the algorithm predictive power, decreasing the log-loss metric (\emph{proportion\_cat}: $p$-value = 0.016, estimate = -0.758; 95\% CI [-1.340; -0.175]; \emph{avg\_cardinality}: $p$-value = 0.007, estimate = -0.511; 95\% CI [-0.859; -0.164]).

\subsection{\emph{Does the type of encoding influence the training duration time?}}
\noindent
\emph{Binary datasets}

We analyzed the effects of the encoding type and the characteristics of categorical features on \emph{training\_duration}. The most parsimonious LMM included an interaction between \emph{encoding} and \emph{high\_cardinality}, their individual effects, and the \emph{proportion\_cat} variable (see Table 7). 
% table 7
\begin{table}[h]
\fontsize{10}{15}\selectfont 
\resizebox{\linewidth}{!}{
\begin{tabular}{|cc|r|r|r|c|}
\hline
\multicolumn{2}{|c|}{Fixed effects} & \multicolumn{1}{c|}{Estimate} & \multicolumn{1}{c|}{95\% Confidence interval} & \multicolumn{1}{c|}{\emph{p}-value} & Significance level \\ \hline
\multicolumn{2}{|c|}{(Intercept)} & 3.661 & {[}3.228; 4.093{]} & \textless{}0.001 & *** \\ \hline
\multicolumn{2}{|c|}{\emph{High cardinality} (binary)} & -0.315 & {[}-1.032; 0.403{]} & 0.396 &  \\ \hline
\multicolumn{2}{|c|}{\emph{Proportion of categorical   features} (continuous)} & -0.796 & {[}-1.405; -0.187{]} & 0.015 & * \\ \hline
\multicolumn{1}{|c|}{\multirow{4}{*}{\begin{tabular}[c]{@{}c@{}}\emph{Encoder}: \\    target encoder vs.\end{tabular}}} & Helmert & 0.297 & {[}-0.019; 0.613{]} & 0.076 & . \\ \cline{2-6} 
\multicolumn{1}{|c|}{} & OHE & 0.189 & {[}-0.057; 0.435{]} & 0.142 &  \\ \cline{2-6} 
\multicolumn{1}{|c|}{} & WOE & -0.004 & {[}-0.011; 0.003{]} & 0.298 &  \\ \cline{2-6} 
\multicolumn{1}{|c|}{} & label & -0.075 & {[}-0.108; -0.042{]} & \textless{}0.001 & *** \\ \hline
\multicolumn{1}{|c|}{\multirow{4}{*}{\emph{High cardinality} *\emph{ Encoder}  (target vs. others)}} & Helmert & 1.408 & {[}0.607; 2.210{]} & 0.002 & ** \\ \cline{2-6} 
\multicolumn{1}{|c|}{} & OHE & 1.143 & {[}0.524; 1.762{]} & 0.001 & *** \\ \cline{2-6} 
\multicolumn{1}{|c|}{} & WOE & 0.001 & {[}-0.017; 0.019{]} & 0.907 &  \\ \cline{2-6} 
\multicolumn{1}{|c|}{} & label & 0.108 & {[}0.024; 0.191{]} & 0.016 & * \\ \hline
\end{tabular}}
\caption{Predicting variability in training duration for binary tasks.}
\end{table}

The most significant effect on the outcome had the \emph{encoding}*\emph{high\_cardinality} interaction term: in \emph{high\_cardinality} datasets, the training duration increased when the categorical features were transformed into long vectors, using Helmert encoding ($p$-value = 0.002, estimate = 1.408; 95\% CI [0.607; 2.210]) and OHE ($p$-value = 0.001, estimate = 1.143; 95\% CI [0.524; 1.762]). We observed a similar pattern, but with a much more modest coefficient, when the categorical features in \emph{high\_cardinality} datasets were encoded into integers, using label encoding ($p$-value = 0.016, estimate = 0.108; 95\% CI [0.024; 0.191]). As an individual factor, however, label encoding significantly decreased the training duration ($p$-value $<$ 0.001, estimate = -0.075; 95\% CI [-0.108; -0.042]); while Helmert encoding was a marginally significant contributor to the increase in training time ($p$-value = 0.076, estimate = -0.075; 95\% CI [-0.019; 0.613]), compared to other types of encoding. The \emph{proportion\_cat} variable had a modest negative effect on training duration ($p$-value = 0.015, estimate = -0.796; 95\% CI [-1.405; -0.187]). This might be explained by the fact that the majority of the base learner estimators in H2O AutoML are tree-based models: branching for continuous variables requires finding appropriate threshold values at every split point, which might significantly increase training time. \newline

\noindent
\emph{Multiclass datasets}
\newline
\indent For this particular LMM analysis we relaxed our initial assumptions for the random intercept, random slope and their correlation for each dataset across the encoding types. Due to non-convergence of the models with and without interaction terms, we only included random intercept and random slope without their correlation. The final model summarized in Table 8 included only main effects.
% table 8
\begin{table}[h]
\fontsize{10}{15}\selectfont 
\resizebox{\linewidth}{!}{
\begin{tabular}{|cc|r|r|r|c|}
\hline
\multicolumn{2}{|c|}{Fixed effects} & \multicolumn{1}{c|}{Estimate} & \multicolumn{1}{c|}{95\% Confidence interval} & \multicolumn{1}{c|}{\emph{p}-value} & Significance level \\ \hline
\multicolumn{2}{|c|}{(Intercept)} & 0.140 & {[}-1.048; 1.328{]} & 0.819 &  \\ \hline
\multicolumn{1}{|c|}{\multirow{4}{*}{\begin{tabular}[c]{@{}c@{}}\emph{Encoder}: \\    target encoder vs.\end{tabular}}} & label & -0.263 & {[}-0.431; -0.095{]} & 0.004 & ** \\ \cline{2-6} 
\multicolumn{1}{|c|}{} & WOE & -0.006 & {[}-0.018; 0.005{]} & 0.299 &  \\ \cline{2-6} 
\multicolumn{1}{|c|}{} & OHE & 0.230 & {[}-0.028; 0.487{]} & 0.089 & . \\ \cline{2-6} 
\multicolumn{1}{|c|}{} & Helmert & 0.342 & {[}0.047; 0.637{]} & 0.029 & * \\ \hline
\multicolumn{2}{|c|}{\emph{Proportion of categorical features} (continuous)} & 2.418 & {[}1.503; 3.333{]} & \textless{}0.001 & *** \\ \hline
\multicolumn{2}{|c|}{\emph{Number of target classes} (integer)} & 1.936 & {[}1.397; 2.475{]} & \textless{}0.001 & *** \\ \hline
\end{tabular}}
\caption{Predicting variability in training duration for multiclass tasks.}
\end{table}

Two predictors showed the most significant effect on the outcome: the number of target classes ($p$-value < 0.001, estimate = 1.936; 95\% CI [1.397; 2.475]) and the proportion of categorical features in a dataset ($p$-value $<$ 0.001, estimate = 2.418; 95\% CI [1.503; 3.333]), indicating that both were associated with significant increases in training durations. The encoding scheme also played a significant role on the training duration: Predictably, label encoding was associated with decreased training time ($p$-value = 0.004, estimate = -0.263; 95\% CI [-0.431; -0.095]); in contrast, Helmert had the opposite effect, increasing the total training duration ($p$-value = 0.029, estimate = 0.342; 95\% CI [0.047; 0.637]); a similar marginally significant trend was observed for the OHE ($p$-value = 0.089, estimate = 0.230; 95\% CI [-0.028; 0.487]). However, some violations on the diagnostic Q-Q plots advise caution against putting too much confidence in marginal trends.

\section{Discussion}
This paper explored the effects of categorical variable encoding for different types of classification tasks in an exhaustive sample from the OpenML database. We fitted LMMs to the data, addressing three research questions. Below we summarize our findings from the analysis. We additionally provide a flowchart, describing some of the optimal pre-processing schemes for categorical data (see Figure 3).
\begin{figure*}[h]
\centering
\includegraphics[width=0.8\textwidth]{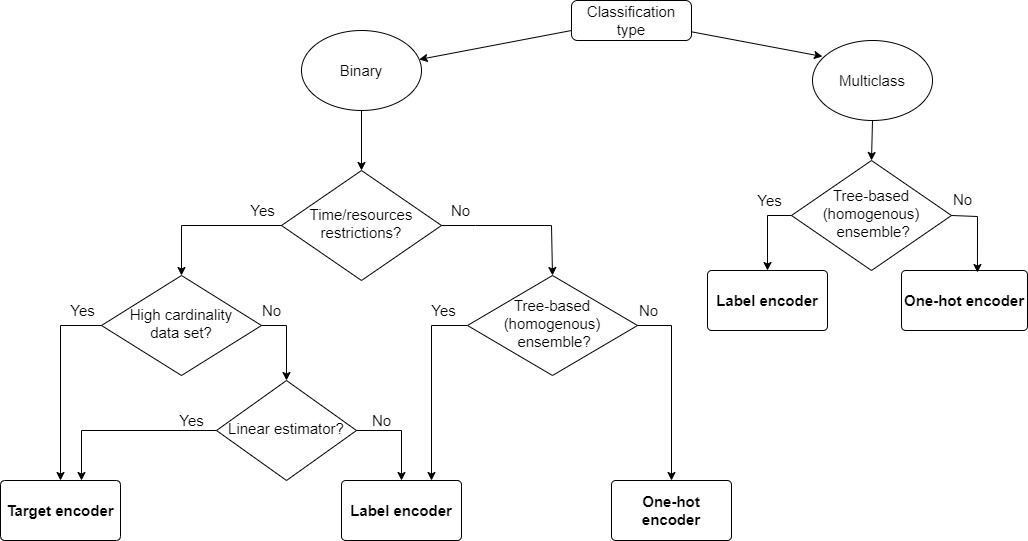}
\caption{Flowchart with recommended options for categorical
data encoding based on our findings from 47 binary and 36 multiclass datasets.}
\end{figure*}
\subsection{Research Question 1: Does the encoding type influence the predictive power of a heterogeneous ensemble?}

In solving real production problems, heterogeneous ensembles typically demonstrate greater predictive power compared to their constituent individual estimators, such as linear or non-linear (tree-based and deep learning) models \citep{Laan2007}. If the encoding effects in base learners were small or inconsistent (for example, relevant for one type of base estimator), we predicted that they might become negligible in a stacked ensemble.

For binary tasks, we did not observe any effects of the encoding type on the stacked ensemble metric, even controlling for variability in characteristics of categorical features (such as their proportion, cardinality, importance for the model, and so on). This confirms that both target-based and target-agnostic encoders could be used interchangeably without any significant effects on the outcome variable.

For multiclass problems, however, we did observe minor beneficial effects of the encoding. However, counter to our predictions, target-agnostic indicator encoders (OHE and Helmert contrast coding) performed better compared to Label encoder and target-based encoders. Possible explanation for the detrimental effects of target-based encoders was overfitting: the outcome variable distributions likely differed across training and validation dataframes, and those differences became more apparent in multiclass problems, particularly when the number of predicted target classes was large. Although this explanation sounds reasonable, it remains unclear, why we did not observe differences between target-based encoders and the baseline label encoder, which is a target-agnostic integer scheme.

We did not observe interactions between the type of encoding and the characteristics of the categorical features. This suggests that our conclusions apply to a wide variety of datasets, regardless of the number/proportion of categorical features, their cardinality, importance, and so on.

We additionally identified general patterns associated with data complexity that might be beneficial for solving data science problems across domains. In binary tasks, increasing the proportion of categorical features negatively affected the predictive power of a stacked ensemble: numerical features seem to contain more useful information for the models. This assumption is supported by the observed positive effects of the average cardinality on the outcome: categorical features with a greater number of levels appeared more informative. However, when cardinality was excessively high (in our sample, the average cardinality over 90 levels across all available categorical features in a dataset), the task became significantly more challenging for the ML algorithms. In multiclass tasks, none of these effects were observable: In addition to the encoding type, the only significant factor was the number of target classes. Predictably, the number of levels in the target variable linearly increased the task complexity for a stacked ensemble.

\subsection*{4.2 Research Question 2: Does the type of encoding influence the predictive power of an individual base estimator? Can we observe linear vs. non-linear effects? }

We hypothesized that, despite possible absence of encoding effects in a heterogeneous ensemble, encoding might influence the performance of its constituent base estimators. In particular, we predicted that target-based encoding schemes would positively influence the predictive power of both linear and non-linear models, given recent evidence of their successful performance \citep{Bourdonnaye2021}, \citep{Hien2020}, \citep{Potdar2017}, \citep{Seca2021}, \citep{Valdez2021}. Also we expected that label encoder would decrease the predictive power of the linear estimator, but not of the tree-based homogeneous ensemble, such as random forest or XGBoost.

\subsection*{4.2.1. Effects of encoding on a linear base learner performance: GLM}

In both binary and multiclass datasets, we found similar patterns. Target-agnostic integer encoding (label encoding) significantly decreased the GLM metric; other target-based and target-agnostic encoders did not significantly differ from one another. We attribute this to how label encoding in the default setting randomly assigns integer values to feature levels, introducing a non-existent ordinal relationship between feature levels. For high cardinality features such approach could be particularly damaging.

Counter to what was shown in earlier studies, target-based encoders did not improve model performance. Regarding other effects, we identified similar patterns to those observed for the stacked ensemble models: binary problems were sensitive to the proportion of categorical features and their cardinality; multiclass problems were only influenced by the number of predicted classes. None of the interaction terms were significant.

\subsection*{4.2.2. Effects of encoding on a non-linear base learner performance: XGBoost}

The encoding scheme did not significantly influence the predictive power of the XGBoost model across binary and multiclass problems. High cardinality of the categorical features decreased model performance, but these effects were mitigated by a modest positive contribution of the average cardinality, which suggested that features with more levels generally contain more useful information for the XGBoost algorithm. Interestingly, we identified that the proportion of categorical features in a dataset had the opposite effect in binary and in multiclass problems. We have discussed above, that in binary problems an increase in the \emph{proportion\_cat} value had a detrimental effect on the metric across ML algorithms. In fact, it was the most significant dependent variable across all LMMs in binary tasks. However, in multiclass problems, we observed a positive effect of this feature on model performance: log-loss was smaller in datasets with a greater proportion of categorical variables. To determine generalizability of these findings, we need to investigate these effects in a greater sample of datasets, particularly for multiclass problems, where the available sample was smaller, and the observed pattern was also much less significant. As before, in multiclass problems, the number of target classes had a highly significant negative effect on model performance.

\subsection*{4.3 Research Question 3: Does the type of encoding influence training efficiency?}

We explored the effects of different types of encoding in terms of model training efficiency. This we estimated by using the total training duration of the H2O AutoML as a dependent variable in LMM, while keeping the CPU resources fixed across all experimental conditions. We predicted that in binary tasks target-agnostic indicator encoders – Helmert encoder and OHE – would require greater training time, particularly with high cardinality data. This assumption is based on the shape of the resulting dataset: wide datasets take longer to converge, and Helmert and OHE expand the feature space proportionally to feature cardinality. This effect would be smaller in multiclass tasks, because target-based encoders substitute each categorical feature with a set of $M-1$ features, where $M$ is the number of target classes. Thus, the differences between target-agnostic indicator encoders (OHE and Helmert) and the target-based encoders (WOE and target encoders) would likely diminish or disappear. Label encoder would likely be the most efficient timewise as it did neither change the shape of the original dataset, nor required calculation of additional statistics.

In general, our predictions were confirmed. In binary datasets, training time was significantly influenced by the type of encoding in interaction with the dataset cardinality: high cardinality datasets encoded with Helmert or OHE required significantly more training time. Label encoder generally led to minor reductions in training duration, but in interaction with \emph{high\_cardinality}, it was slightly less efficient than target encoder. In multiclass datasets, as we predicted, OHE and Helmert encoder were only marginally worse compared to target-based encoders. Label encoder, as expected, was the most efficient timewise.

\section{Limitations and Future Directions}

Our study has several limitations that could be overcome in future projects. First, although we searched exhaustively the entire OpenML repository, the resulting sample of datasets for multiclass problems was modest, with only 36 unique data frames included in the analysis. Given the diversity of the real world problems, adding more datasets would allow checking generalizability of our findings. Second, our study focused only on three general types of encoding schemes: target-based encoding (WOE and target encoder), target-agnostic indicator encoding (OHE and Helmert encoder) and target-agnostic integer encoding (label encoder), which we used as a baseline. Other types of encoding are likely more suitable for specific types of categorical data, such as textual string information. These include bag of n-grams, word embeddings or other less common, but efficient methods, such as gamma-poisson matrix factorization or min-hash encoding \citep{Cerda2019}. However, some of these encoding schemes (e.g., embeddings) require deep learning models as estimators. In addition, OpenML does not offer sufficient number of datasets of the suitable sort to study these effects in detail; we therefore, leave this out for future research.

\section{Conclusions}
 
This paper explored the effects of encoding on model accuracy and training duration classification problems. For the first time, such study took into account possible interactions between the encoding type and dataset features: the characteristics of categorical variables, their proportion in a dataset, and individual differences between tasks (we assumed random slope and intercept for each problem). Overall, the encoding type did not play a significant role across the majority of estimators and tasks, counter to what has been reported before[5-8, 18]. Importantly, the interactions terms between the type of encoding and the dataset characteristics were non-significant: our findings about the encoding effects were consistent across datasets, regardless of the number of categorical features, their proportion, cardinality and even their importance for individual linear and non-linear estimators. This suggests that our observations are generalizable to a wide variety of problems across domains.

For a stacked ensemble model in binary tasks we identified a positive trend towards significance for OHE, but no difference for other conditions; in multiclass problems, we observed beneficial effect of target-agnostic indicator encoding (OHE and Helmert encoder), but the effect was modest and requires further investigation in a larger sample of datasets. Surprisingly, none of the encoding effects were observed for the XGBoost algorithm; similarly, for the GLM, we identified only the predictable negative effect of label encoding, which introduced a scale relationship between feature levels, which was absent in real data. 
In terms of efficiency, we confirmed our predictions for both problem types. In binary tasks, OHE and Helmert lead to significantly greater training time for high cardinality datasets, suggesting better suitability of alternative encoders under resource restrictions. In contrast, label encoder was associated with decreased training time overall; interestingly, in high cardinality data, using target encoder was more efficient. In multiclass tasks, Helmert/OHE encoding were only marginally worse compared to WOE/target encoding: target-based schemes increase the feature space proportionally to the number of target classes. Transforming multiclass data with label encoder lead to maximal reductions in training time.

In terms of efficiency, we confirmed our predictions for both types of problems. In binary tasks, OHE and Helmert lead to significantly greater training time for high cardinality datasets. This suggest that it is not advisable to use these encoders under resource restrictions. In contrast, label encoder was associated with decreased training time overall; interestingly, in high cardinality data, using target encoder was more efficient. In multiclass tasks, target-agnostic indicator encoding were only marginally worse compared to target-based encoding; this is probably because with target and WOE encoders the size of the transformed feature space depends on the number of target classes, and it is often comparable to the one produced by OHE and Helmert encoders. Transforming multiclass data with label encoder lead to maximal reductions in training time.

%%\label{}
%\lipsum[1]

%\subsection{Subsection title}

%\begin{figure}
%	\centering 
%	\includegraphics[width=0.4\textwidth, angle=-90]{ASCOM_journal_cover.pdf}	
%	\caption{Astronomy \& Computing journal cover} 
%	\label{fig_mom0}%
%\end{figure}

%%\label{}
%\lipsum[2]

%\subsection{Subsection title}
%\lipsum[3]

%\begin{table}
%\begin{tabular}{l c c c} 
% \hline
% Source & RA (J2000) & DEC (J2000) & $V_{\rm sys}$ \\ 
%        & [h,m,s]    & [o,','']    & \kms          \\
% \hline
% NGC\,253 & 	00:47:33.120 & -25:17:17.59 & $235 \pm 1$ \\ 
% M\,82 & 09:55:52.725, & +69:40:45.78 & $269 \pm 2$ 	 \\ 
% \hline
%\end{tabular}
%\caption{Random table with galaxies coordinates and velocities, Number the tables consecutively in}
%\label{Table1}
%\end{table}

%\section{Discussion}
%%\label{}
%\lipsum[4]

%\section{Summary and conclusions}
%%\label{}
%\lipsum[1-4]

\section*{Acknowledgements}
We are very grateful to our colleagues from the ML team for their feedback and suggestions on the project, and particularly to Alexey Ratushnii and Dr. Evgenii Vityaev for their helpful comments on the Manuscript.

%\appendix

%\section{Appendix title 1}

%\section{Appendix title 2}


%\bibliographystyle{elsarticle-harv}
\nocite{*}
\bibliographystyle{unsrt}
\setcitestyle{numbers}
\bibliography{main}



%% For example:

%% \bibitem[Aladro et al.(2015)]{Aladro15} Aladro, R., Martín, S., Riquelme, D., et al. 2015, \aas, 579, A101

%%\end{thebibliography}

\end{document}